%% file: paper.tex
\documentclass[10pt,twocolumn,letterpaper]{article}

\usepackage{cvpr}
\usepackage{times}
\usepackage{epsfig}
\usepackage{graphicx}
\usepackage{amsmath}
\usepackage{amssymb}

\usepackage{float}
\usepackage{enumitem}
\usepackage{booktabs}

\cvprfinalcopy 

\ifcvprfinal\fi
\pdfoutput=1

\ifcvprfinal\fi
\begin{document}

\title{Learning Transferable Architectures for Scalable Image Recognition}

\author{Barret Zoph\\
Google Brain\\
{\tt\small barretzoph@google.com}
\and
Vijay Vasudevan\\
Google Brain\\
{\tt\small vrv@google.com}
\and
Jonathon Shlens\\
Google Brain\\
{\tt\small shlens@google.com}
\and
Quoc V. Le\\
Google Brain\\
{\tt\small qvl@google.com}
}

\maketitle

\input{abstract.tex}
\input{intro.tex}
\input{related.tex}

\input{methods.tex}
\input{experiments.tex}
\input{conclusions.tex}

{\small
\bibliographystyle{ieee}
\bibliography{paper}
}

\clearpage
\input{appendix.tex}

\end{document}

%% file: abstract.tex
\begin{abstract}

 Developing neural network image classification models often requires significant architecture engineering.  In this paper, we study a method to learn the model architectures directly on the dataset of interest. As this approach is expensive when the dataset is large, we propose to search for an architectural building block on a small dataset and then transfer the block to a larger dataset. The key contribution of this work is the design of a new search space (which we call the ``NASNet search space") which enables transferability.
 In our experiments, we search for the best convolutional layer (or ``cell") on the CIFAR-10 dataset and then apply this cell to the ImageNet dataset by stacking together more copies of this cell, each with their own parameters to design a convolutional architecture, which we name a ``NASNet architecture". We also introduce a new regularization technique called ScheduledDropPath that significantly improves generalization in the NASNet models. On CIFAR-10 itself, a NASNet found by our method achieves 2.4\% error rate, which is state-of-the-art. Although the cell is not searched for directly on ImageNet, a NASNet constructed from the best cell achieves, among the published works, state-of-the-art accuracy of 82.7\% top-1 and 96.2\% top-5 on ImageNet. Our model is 1.2\% better in top-1 accuracy than the best human-invented architectures while having 9 billion fewer FLOPS -- a reduction of 28\% in computational demand from the previous state-of-the-art model.  When evaluated at different levels of computational cost, accuracies of NASNets exceed those of the state-of-the-art human-designed models. For instance, a small version of NASNet also achieves 74\% top-1 accuracy, which is 3.1\% better than equivalently-sized, state-of-the-art models for mobile platforms.
 Finally, the image features learned from image classification are generically useful and can be transferred to other computer vision problems. On the task of object detection, the learned features by NASNet used with the Faster-RCNN framework surpass state-of-the-art by 4.0\% achieving 43.1\% mAP on the COCO dataset.
  
  \end{abstract}

%% file: intro.tex
\section{Introduction}

Developing neural network image classification models often requires significant \emph{architecture engineering}. Starting from the seminal work of~\cite{krizhevsky2012imagenet} on using convolutional architectures \cite{fukushima1982neocognitron,lecun1998gradient} for ImageNet~\cite{deng2009imagenet} classification, successive advancements through architecture engineering have achieved impressive results~\cite{simonyan2014very,szegedy2015going,he2015deep,szegedy2016rethinking,szegedy2016inception,xie2016aggregated}. 


In this paper, we study a new paradigm of designing convolutional architectures and describe a scalable method to optimize convolutional architectures on a dataset of interest, for instance the ImageNet classification dataset. 
Our approach is inspired by the recently proposed Neural Architecture Search (NAS) framework~\cite{zoph2017neural}, which uses a reinforcement learning search method to optimize architecture configurations. Applying NAS, or any other search methods, directly to a large dataset, such as the ImageNet dataset, is however computationally expensive. We therefore propose to search for a good architecture on a proxy dataset, for example the smaller CIFAR-10 dataset, and then transfer the learned architecture to ImageNet. We achieve this transferrability by designing a search space (which we call ``the NASNet search space") so that the complexity of the architecture is independent of the depth of the network and the size of input images. More concretely, all convolutional networks in our search space are composed of convolutional layers (or ``cells") with identical structure but different weights. Searching for the best convolutional architectures is therefore reduced to searching for the best cell structure. Searching for the best cell structure has two main benefits: it is much faster than searching for an entire network architecture and the cell itself is more likely to generalize to other problems. 
In our experiments, this approach significantly accelerates the search for the best architectures using CIFAR-10 by a factor of $7\times$ and learns architectures that successfully transfer to ImageNet.

Our main result is that the best architecture found on CIFAR-10, called NASNet,
achieves state-of-the-art accuracy when transferred to ImageNet classification without much modification. On ImageNet, NASNet achieves, among the published works, state-of-the-art accuracy of 82.7\% top-1 and 96.2\% top-5. This result amounts to a 1.2\% improvement in top-1 accuracy than the best human-invented architectures while having 9 billion fewer FLOPS. On CIFAR-10 itself, NASNet achieves 2.4\% error rate, which is also state-of-the-art.



  Additionally, by simply varying the number of the convolutional cells and number of filters in the convolutional cells, we can create different versions of NASNets with different computational demands. Thanks to this property of the cells, we can generate a family of models that achieve accuracies superior to all human-invented models at equivalent or smaller computational budgets \cite{szegedy2016rethinking,BatchNorm}. Notably, the smallest version of NASNet achieves 74.0\% top-1 accuracy on ImageNet, which is 3.1\% better than previously engineered architectures targeted towards mobile and embedded vision tasks \cite{howard2017mobilenets,shufflenet}.

Finally, we show that the image features learned by NASNets are generically useful and transfer to other computer vision problems. In our experiments, the features learned by NASNets from ImageNet classification can be combined with the Faster-RCNN framework~\cite{faster_rcnn} to achieve state-of-the-art on COCO object detection task for both the largest as well as mobile-optimized models. Our largest NASNet model achieves 43.1\% mAP, which is 4\% better than previous state-of-the-art. 



%% file: related.tex
\section{Related Work}

The proposed method is related to previous work in hyperparameter optimization~\cite{pinto2009high,bergstra2011algorithms,bergstra2012random,snoek2012practical,snoek2015scalable,bergstra2013making,mendoza2016towards} -- especially recent approaches in designing architectures such as Neural Fabrics~\cite{saxena2016convolutional}, DiffRNN~\cite{miconi2016}, MetaQNN~\cite{baker2016designing} and DeepArchitect~\cite{negrinho2017}.  A more flexible class of methods for designing architecture  is evolutionary algorithms \cite{wierstra2005modeling,floreano2008neuroevolution,stanley2009hypercube,jozefowicz2015empirical,real2017large,miikkulainen2017evolving,xie17}, yet they have not had as much success at large scale. 
Xie and Yuille \cite{xie17} also transferred learned architectures from CIFAR-10 to ImageNet but performance of these models (top-1 accuracy 72.1\%) are notably below previous state-of-the-art (Table \ref{tab:imagenet}).

The concept of having one neural network interact with a second neural network to aid the learning process, or learning to learn or meta-learning \cite{hochreiter2001learning,schaul10} has attracted much attention in recent years~\cite{andrychowicz2016learning,wang2016learning,duan2016rl,ha2016hypernetworks,li2017learning,ravi2017optimization,finn2017model}. Most of these approaches have not been scaled to large problems like ImageNet. An exception is the recent work focused on learning an optimizer for ImageNet classification that achieved notable improvements \cite{wichrowska2017learned}.

The design of our search space took much inspiration from LSTMs~\cite{lstm}, and Neural Architecture Search Cell~\cite{zoph2017neural}. The modular structure of the convolutional cell is also related to previous methods on ImageNet such as VGG \cite{simonyan2014very}, Inception~\cite{szegedy2015going,szegedy2016rethinking,szegedy2016inception}, ResNet/ResNext~\cite{he2015deep,xie2016aggregated}, and Xception/MobileNet \cite{chollet2016xception, howard2017mobilenets}.

%% file: methods.tex
\section{Method}
\label{sec:search_space}



\begin{figure}[h!]
\begin{center}
\includegraphics[width=1.0\columnwidth]{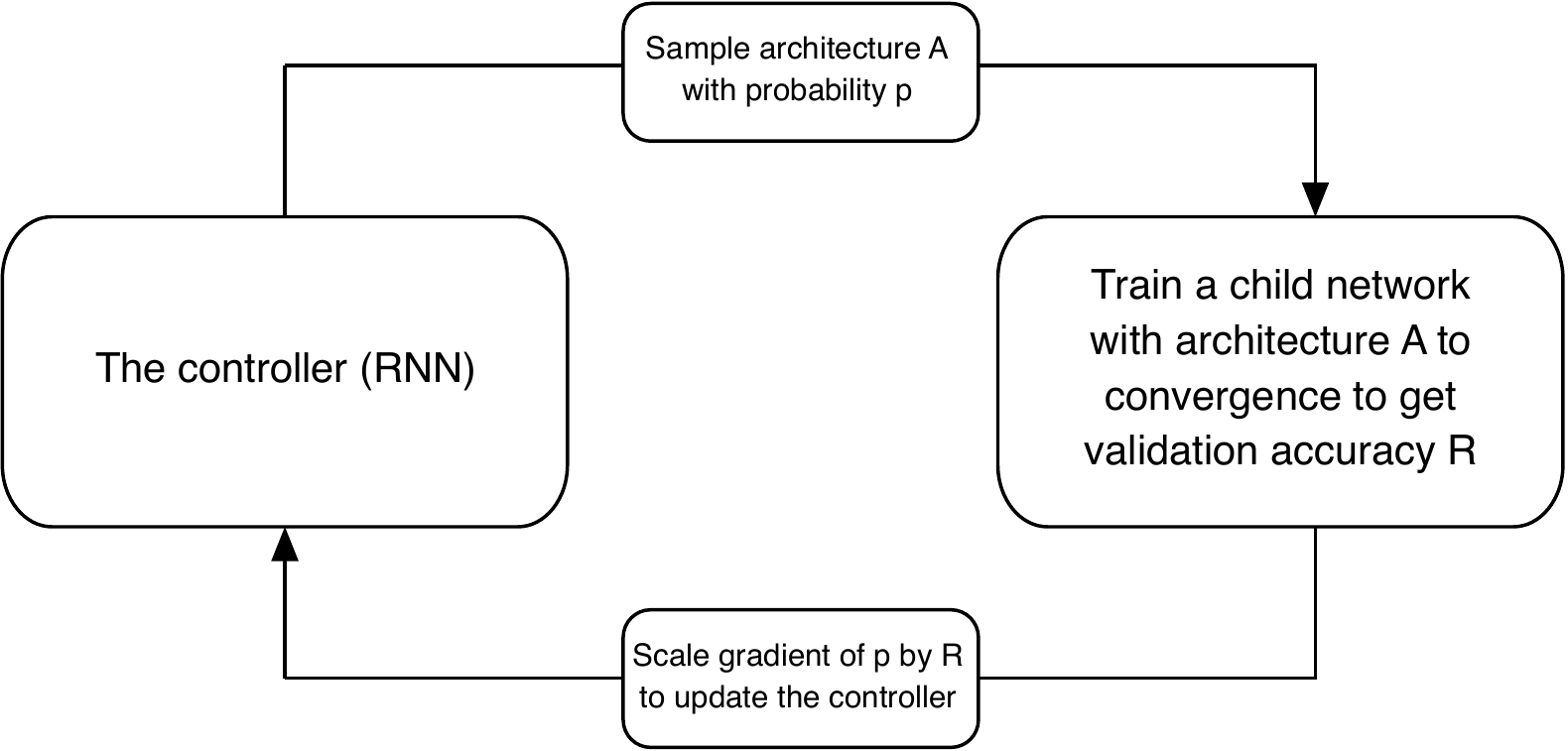}
\caption{Overview of Neural Architecture Search \cite{zoph2017neural}. A controller RNN predicts architecture $A$ from a search space with probability $p$. A child network with architecture $A$ is trained to convergence achieving accuracy $R$. Scale the gradients of $p$ by $R$ to update the RNN controller.}
\label{figure:controller}
\end{center}
\end{figure}

Our work makes use of search methods to find good convolutional architectures on a dataset of interest. The main search method we use in this work is the Neural Architecture Search (NAS) framework proposed by~\cite{zoph2017neural}. In NAS, a controller recurrent neural network (RNN) samples child networks with different architectures. The child networks are trained to convergence to obtain some accuracy on a held-out validation set. The resulting accuracies are used to update the controller so that the controller will generate better architectures over time. The controller weights are updated with policy gradient (see Figure~\ref{figure:controller}). 

The main contribution of this work is the design of a novel search space, such that the best architecture found on the CIFAR-10 dataset would scale to larger, higher-resolution image datasets across a range of computational settings.
We name this search space \emph{the NASNet search space} as it gives rise to NASNet, the best architecture found in our experiments. One inspiration for the NASNet search space is the realization that architecture engineering with CNNs often identifies 
repeated motifs consisting of combinations of convolutional
filter banks, nonlinearities and a prudent selection of
connections to achieve state-of-the-art results (such as the repeated modules present in the Inception and ResNet models \cite{szegedy2015going,he2015deep,szegedy2016rethinking,szegedy2016inception}). These observations suggest that it may be possible for the controller RNN to predict a generic {\it convolutional cell} expressed in terms of these motifs. This cell can then be stacked in series to handle inputs of arbitrary spatial dimensions and filter depth.

In our approach, the overall architectures of the convolutional nets are manually predetermined. They are composed of convolutional cells repeated many times where each convolutional cell has the same architecture, but different weights. To easily build scalable architectures for images of any size, we need two types of convolutional cells to serve two main functions when taking in a feature map as input: (1) convolutional cells that return a feature map of the same dimension,  and (2) convolutional cells that return a feature map where the feature map height and width is reduced by a factor of two. We name the first type and second type of convolutional cells  \emph{Normal Cell} and \emph{Reduction Cell} respectively. For the Reduction Cell, we make the initial operation applied to the cell's inputs have a stride of two to reduce the height and width. All of our operations that we consider for building our convolutional cells have an option of striding.

\begin{figure}[t!]
\begin{center}
\includegraphics[width=0.7\columnwidth]{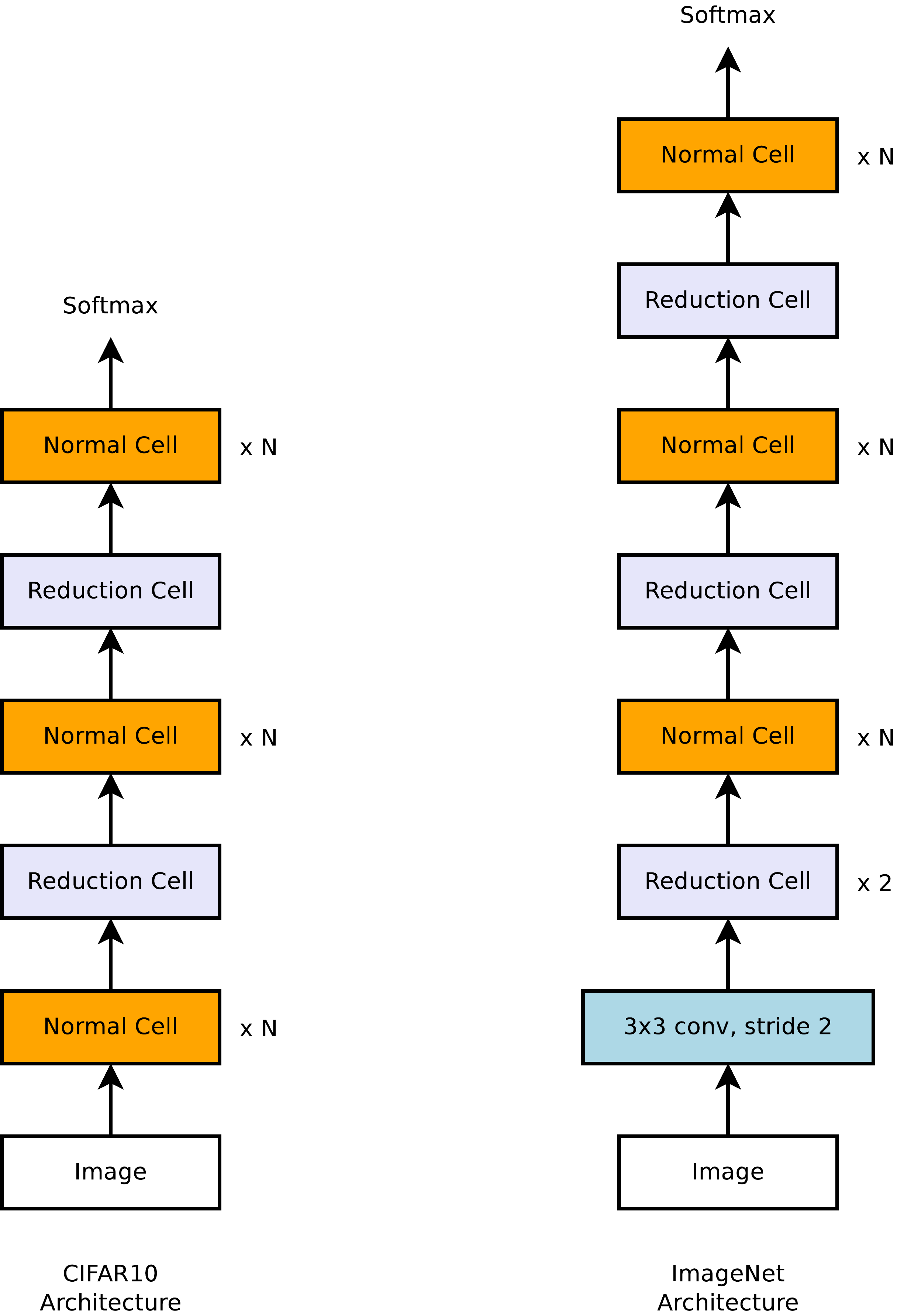}
\caption{Scalable architectures for image classification consist of two repeated motifs termed {\it Normal Cell} and {\it Reduction Cell}. This diagram highlights the model architecture for CIFAR-10 and  ImageNet. The choice for the number of times the Normal Cells that gets stacked between reduction cells, $N$, can vary in our experiments.}
\label{figure:mainnet}
\end{center}
\end{figure}

Figure~\ref{figure:mainnet} shows our placement of Normal and Reduction Cells for CIFAR-10 and ImageNet. Note on ImageNet we have more Reduction Cells, since the incoming image size is 299x299 compared to 32x32 for CIFAR. The Reduction and Normal Cell could have the same architecture, but we empirically found it beneficial to learn two separate architectures.
We use a common heuristic to double the number of filters in the output whenever the spatial activation size is reduced in order to maintain roughly constant hidden state dimension \cite{krizhevsky2012imagenet,simonyan2014very}.
Importantly,
much like Inception and ResNet models \cite{szegedy2015going,he2015deep,szegedy2016rethinking,szegedy2016inception},
we consider the number of motif repetitions $N$ and the number of initial convolutional filters as free parameters that we tailor to the scale of an image classification problem.





What varies in the convolutional nets is the structures of the Normal and Reduction Cells, which are searched by the controller RNN. 
The structures of the cells can be searched within a search space defined as follows (see Appendix, Figure \ref{figure:nas-search-space} for schematic). In our search space, each cell receives as input two initial hidden states $h_i$ and $h_{i-1}$ which are the outputs of two cells in previous two lower layers or the input image. The controller RNN recursively predicts the rest of the structure of the convolutional cell, given these two initial hidden states (Figure~\ref{figure:cell}). The predictions of the controller for each cell are grouped into $B$ blocks, where each block has 5 prediction steps made by 5 distinct softmax classifiers corresponding to discrete choices of the elements of a block:


\begin{figure*}[h!]
\begin{center}
\includegraphics[width=0.7\linewidth]{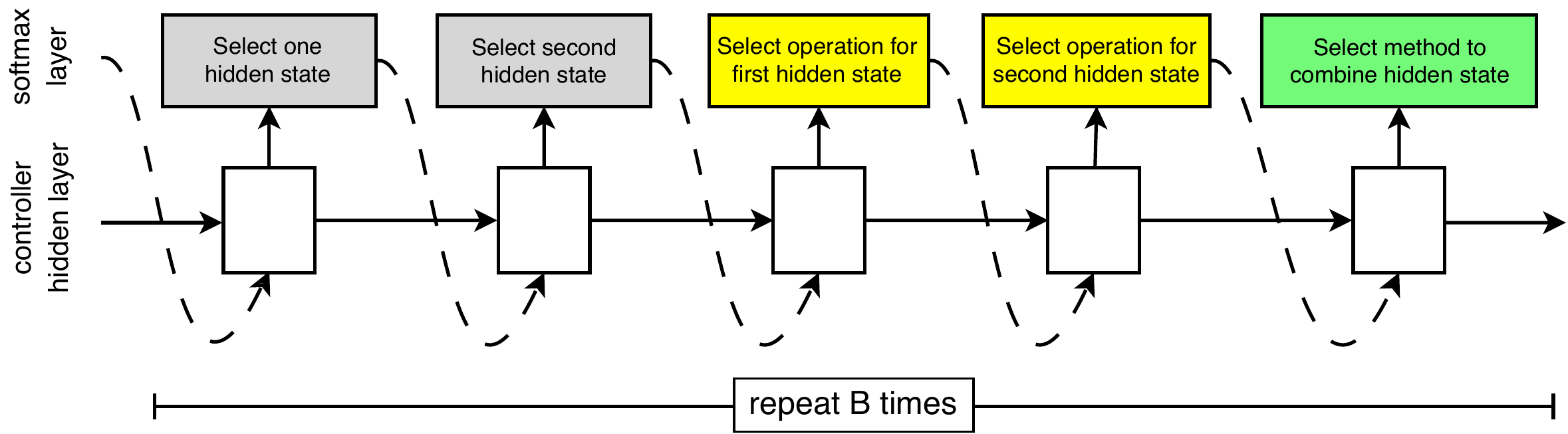}
\hfill
\includegraphics[width=0.2\linewidth]{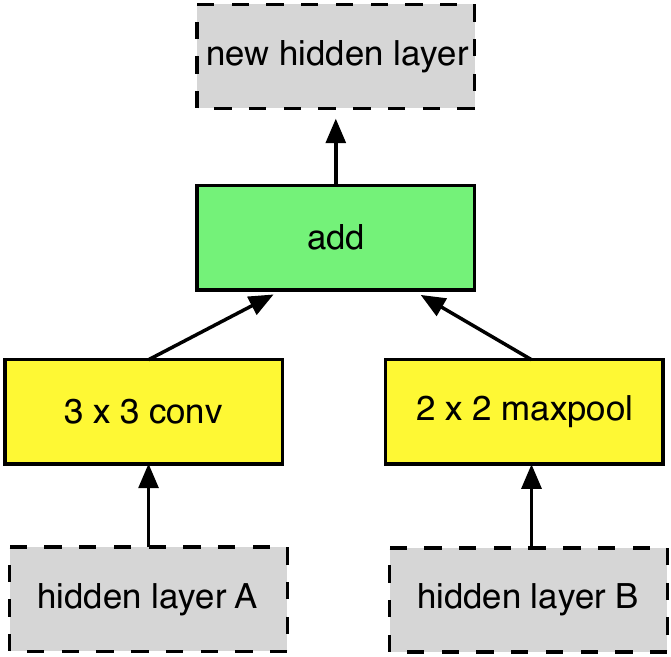}
\caption{Controller model architecture for recursively constructing one block of a convolutional cell. Each block requires selecting 5 discrete parameters, each of which corresponds to the output of a softmax layer. Example constructed block shown on right. A convolutional cell contains $B$ blocks, hence the controller contains $5B$ softmax layers for
predicting the architecture of a convolutional cell. In our experiments, the number of blocks $B$ is 5.}
\label{figure:cell}
\end{center}
\end{figure*}

\begin{enumerate}[itemindent=0.3cm,label={\bf Step \arabic*.}]
\footnotesize
\item Select a hidden state from $h_i, h_{i-1}$ or from the set of hidden states created in previous blocks.
\item Select a second hidden state from the same options as in Step 1.
\item Select an operation to apply to the hidden state selected in Step 1. 
\item Select an operation to apply to the hidden state selected in Step 2.
\item Select a method to combine the outputs of Step 3 and 4 to create a new hidden state.
\end{enumerate}
The algorithm appends the newly-created hidden state to the set of existing hidden states as a potential input in subsequent blocks. The controller RNN repeats the above 5 prediction steps $B$ times corresponding to the $B$ blocks in a convolutional cell. In our experiments, selecting $B=5$ provides good results, although we have not exhaustively searched this space due to computational limitations. 

In steps 3 and 4, the controller RNN selects an operation to apply to the hidden states. We collected the following set of operations based on their prevalence in the CNN literature:

\begin{table}[H]
\footnotesize
\setlength{\tabcolsep}{0.5em} 
\centering
\def\arraystretch{1.0}
\begin{tabular}{ll}
$\bullet\;$ identity & $\bullet\;$ 1x3 then 3x1 convolution \\
$\bullet\;$ 1x7 then 7x1 convolution & $\bullet\;$ 3x3 dilated convolution \\
$\bullet\;$ 3x3 average pooling & $\bullet\;$ 3x3 max pooling \\
$\bullet\;$ 5x5 max pooling & $\bullet\;$ 7x7 max pooling \\
$\bullet\;$ 1x1 convolution & $\bullet\;$ 3x3 convolution \\
$\bullet\;$ 3x3 depthwise-separable conv & $\bullet\;$ 5x5 depthwise-seperable conv \\
$\bullet\;$ 7x7 depthwise-separable conv \\
\end{tabular}
\end{table}

In step 5 the controller RNN selects a method to combine the two hidden states, either (1) element-wise addition between two hidden states or (2) concatenation between two hidden states along the filter dimension.
Finally, all of the unused hidden states generated in the convolutional cell are concatenated together in depth to provide the final cell output.

To allow the controller RNN to predict both Normal Cell and Reduction Cell, we simply make the controller have $2\times 5B$ predictions in total, where the first $5B$ predictions are for the Normal Cell and the second $5B$ predictions are for the Reduction Cell.

Finally, our work makes use of the reinforcement learning proposal in NAS~\cite{zoph2017neural}; however, it is also possible to use random search to search for architectures in the NASNet search space. In random search, instead of sampling the decisions from the softmax classifiers in the controller RNN, we can sample the decisions from the uniform distribution. In our experiments, we find that random search is slightly worse than reinforcement learning on the CIFAR-10 dataset. Although there is value in using reinforcement learning, the gap is smaller than what is found in the original work of~\cite{zoph2017neural}. This result suggests that 1) the NASNet search space is  well-constructed such that random search can perform reasonably well and 2) random search is a difficult baseline to beat. We will compare reinforcement learning against random search in Section~\ref{sec:random_search}.




%% file: experiments.tex
\section{Experiments and Results}

In this section, we describe our experiments with the method described above to learn convolutional cells. In summary, all architecture searches are performed using the CIFAR-10 classification task \cite{krizhevsky2009learning}. The controller RNN was trained using Proximal Policy Optimization (PPO) \cite{SchulmanWDRK17} by employing a global workqueue system for generating a pool of child networks controlled by the RNN. In our experiments, the pool of workers in the workqueue consisted of 500 GPUs. 

The result of this search process over 4 days yields several candidate convolutional cells. We note that this search procedure is almost $7\times$ faster than previous approaches \cite{zoph2017neural} that took 28 days.\footnote{In particular, we note that previous architecture search \cite{zoph2017neural} used 800 GPUs for 28 days resulting in 22,400 GPU-hours. The method in this paper uses 500 GPUs across 4 days resulting in 2,000 GPU-hours. The former effort used Nvidia K40 GPUs, whereas the current efforts used faster NVidia P100s. Discounting the fact that the we use faster hardware, we estimate that the current procedure is roughly about $7\times$ more efficient.} Additionally, we demonstrate below that the resulting architecture is superior in accuracy.

\begin{figure*}[h!]
\begin{center}
\includegraphics[width=0.7\linewidth]{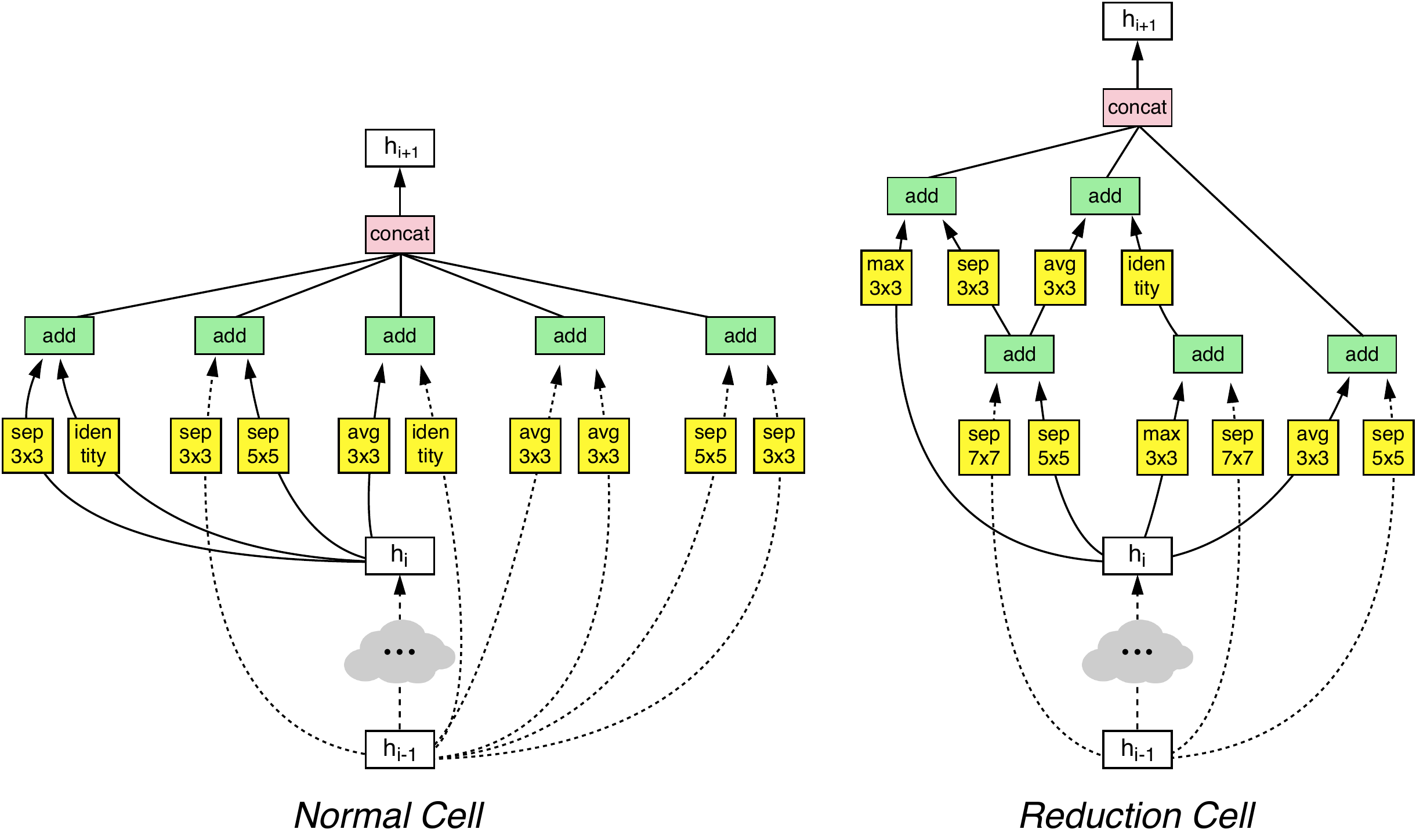}
\caption{Architecture of the best convolutional cells (NASNet-A) with $B=5$ blocks identified with CIFAR-10 . The input (white) is the hidden state from previous activations (or input image). The output (pink) is the result of a concatenation operation across all resulting branches.
Each convolutional cell is the result of $B$ blocks.
A single block is corresponds to two primitive operations (yellow) and a combination operation (green). Note that colors correspond to operations in Figure \ref{figure:cell}. 
}
\label{figure:cell_structure}
\end{center}
\end{figure*}

Figure \ref{figure:cell_structure} shows a diagram of the top performing Normal Cell and Reduction Cell. Note the prevalence of separable convolutions and the number of branches compared with competing architectures \cite{simonyan2014very,szegedy2015going,he2015deep,szegedy2016rethinking,szegedy2016inception}. Subsequent experiments focus on this convolutional cell architecture, although we examine the efficacy of other, top-ranked  convolutional cells in ImageNet experiments (described in Appendix \ref{sec:othernasnet}) and report their results as well. We call the three networks constructed from the best three searches \emph{NASNet-A}, \emph{NASNet-B} and \emph{NASNet-C}.

We demonstrate the utility of the convolutional cells by employing this learned architecture on CIFAR-10 and a family of ImageNet classification tasks. The latter family of tasks is explored across a few orders of magnitude in computational budget.
After having learned the convolutional cells, several hyper-parameters may be explored to build a final network for a given task: (1) the number of cell repeats $N$ and (2) the number of filters in the initial convolutional cell. After selecting the number of initial filters, we use a common heuristic to double the number of filters whenever the stride is 2.
Finally, we define a simple notation, e.g.,  $4$ @ $64$, to indicate these two parameters in all networks, where $4$ and $64$ indicate the number of cell repeats and the number of filters in the penultimate layer of the network, respectively.

For complete details of of the architecture learning algorithm and the controller system, please refer to  Appendix \ref{section:appendix-details}. Importantly, when training NASNets, we discovered \emph{ScheduledDropPath}, a modified version of DropPath~\cite{larsson2016fractalnet}, to be an effective regularization method for NASNet. In DropPath~\cite{larsson2016fractalnet}, each path in the cell is stochastically dropped with some fixed probability during training.  In our modified version, ScheduledDropPath, each path in the cell is dropped out with a probability that is \emph{linearly increased over the course of training}. We find that DropPath does not work well for NASNets, while ScheduledDropPath  significantly improves the final performance of NASNets in both CIFAR and ImageNet experiments.

\subsection{Results on CIFAR-10 Image Classification}


For the task of image classification with CIFAR-10, we set $N=4$ or 6 (Figure \ref{figure:mainnet}). The test accuracies of the best architectures are reported in Table~\ref{tab:cifar10} along with other state-of-the-art models. As can be seen from the Table, a large NASNet-A model with cutout data augmentation~\cite{devries2017improved} achieves a state-of-the-art error rate of 2.40\% (averaged across 5 runs), which is slightly better than the previous best record of 2.56\% by \cite{devries2017improved}. The best single run from our model achieves 2.19\% error rate.


\begin{table*}[h!]
\centering
\small
\begin{tabular}{l|cc|c}
\toprule
\multicolumn{1}{c|}{\bf model} & \multicolumn{1}{l}{\bf depth} & \multicolumn{1}{l|}{\bf \# params}  & \bf error rate (\%)  \\ \midrule
DenseNet $(L=40, k=12)$ \cite{Huang2016Densely} & 40 & 1.0M &  5.24  \\
DenseNet$(L=100, k=12)$ \cite{Huang2016Densely} & 100 & 7.0M &  4.10  \\
DenseNet $(L=100, k=24)$ \cite{Huang2016Densely} & 100 & 27.2M &  3.74  \\ 
DenseNet-BC $(L=100, k=40)$ \cite{Huang2016Densely} & 190 & 25.6M &  3.46  \\
\midrule
Shake-Shake 26 2x32d \cite{gastaldi17shakeshake} & 26 & 2.9M & 3.55 \\
Shake-Shake 26 2x96d \cite{gastaldi17shakeshake} & 26 & 26.2M & 2.86 \\
Shake-Shake 26 2x96d + cutout \cite{devries2017improved} & 26 & 26.2M & 2.56 \\
\midrule
NAS v3 \cite{zoph2017neural}& 39 & 7.1M & 4.47  \\
NAS v3 \cite{zoph2017neural}& 39 & 37.4M & 3.65 \\
\midrule
NASNet-A \;(6 @ 768) & - & 3.3M & 3.41 \\
NASNet-A \;(6 @ 768) + cutout & - & 3.3M & 2.65 \\
NASNet-A \;(7 @ 2304) & - & 27.6M & 2.97 \\
NASNet-A \;(7 @ 2304) + cutout & - & 27.6M & 2.40 \\
NASNet-B \;(4 @ 1152) & - & 2.6M & 3.73 \\
NASNet-C \;(4 @ 640) & - & 3.1M & 3.59 \\
\bottomrule
\end{tabular}
\vspace{0.2cm}
\caption{Performance of Neural Architecture Search and other state-of-the-art models on CIFAR-10. All results for NASNet are the mean accuracy across 5 runs.}
\label{tab:cifar10}
\end{table*}





\subsection{Results on ImageNet Image Classification}

We performed several sets of experiments on ImageNet with the best convolutional cells learned from CIFAR-10.
We emphasize that we merely transfer the architectures from CIFAR-10 but train all ImageNet models weights from scratch.

\begin{figure*}[h!]
\begin{center}
\includegraphics[width=0.45\linewidth]{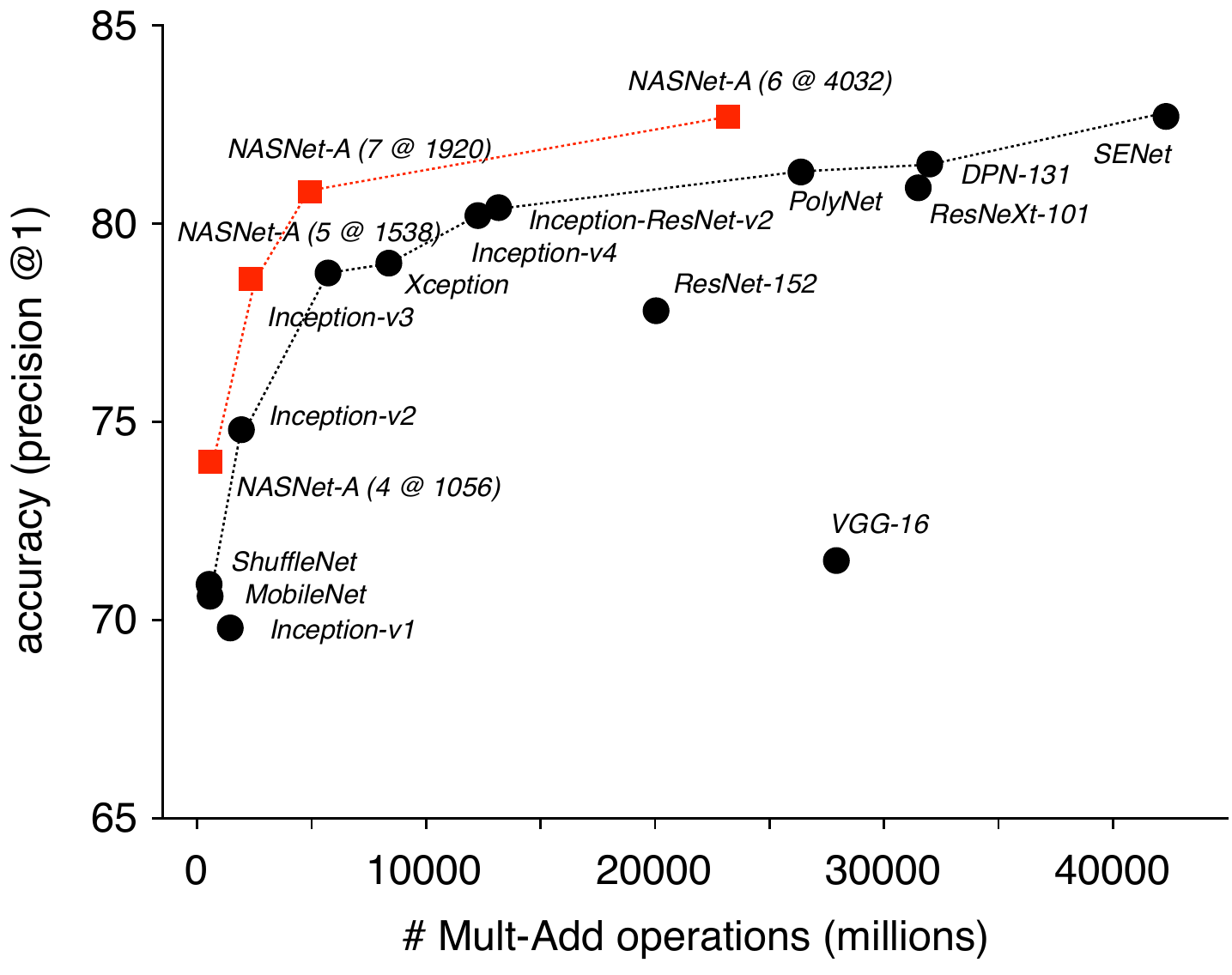}
\hfill
\includegraphics[width=0.45\linewidth]{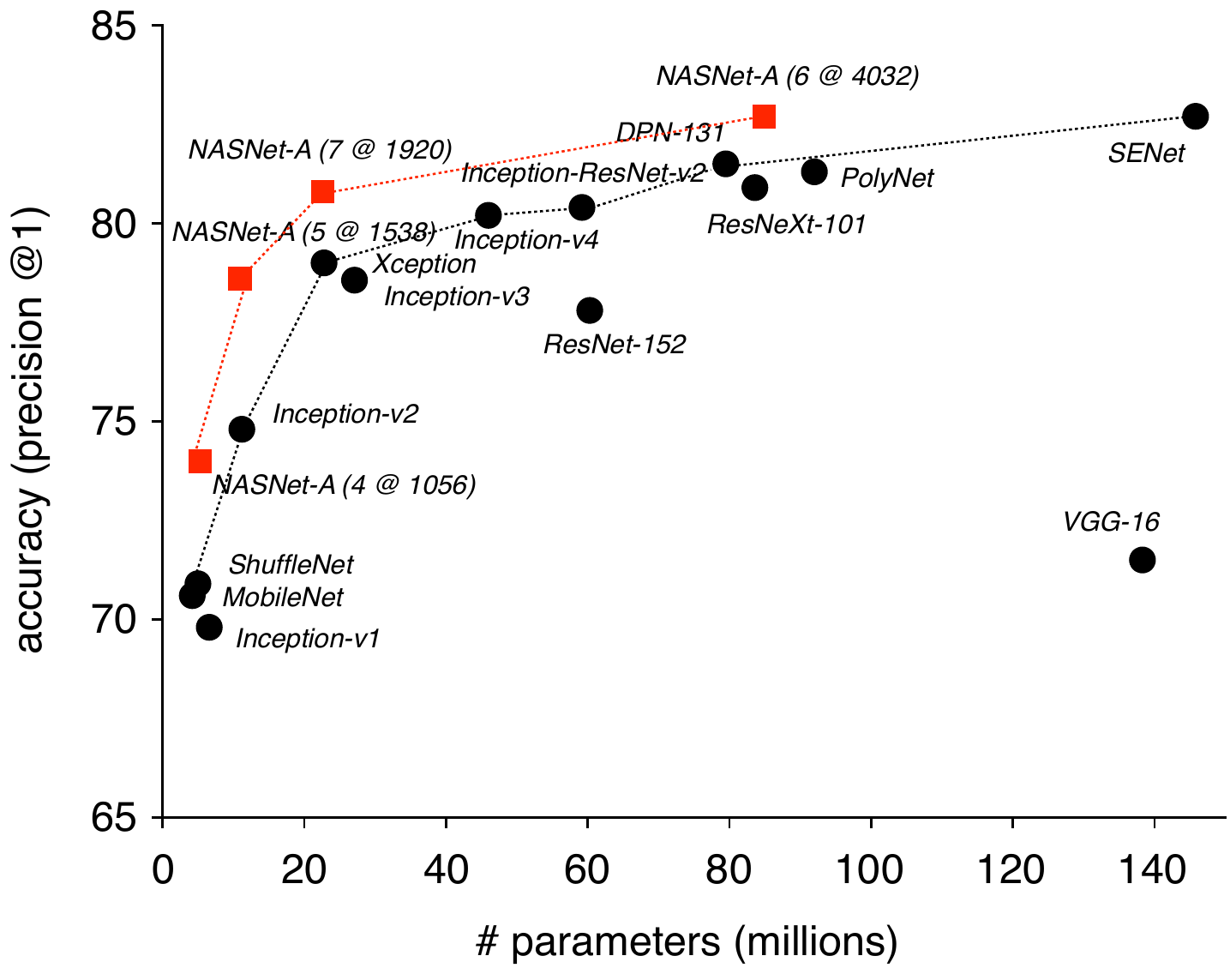}
\caption{Accuracy versus computational demand (left) and number of parameters (right) across top performing published CNN architectures on ImageNet 2012 ILSVRC challenge prediction task. Computational demand is measured in the number of floating-point multiply-add operations to process a single image. Black circles indicate previously published results and red squares highlight our proposed models.}
\label{figure:flops_vs_accuracy}
\end{center}
\end{figure*}

Results are summarized in Table \ref{tab:imagenet} and \ref{tab:imagenet-constrained} and Figure \ref{figure:flops_vs_accuracy}. In the first set of experiments, we train several image classification systems operating on 299x299 or 331x331 resolution images with different experiments scaled in computational demand to create models that are roughly on par in computational cost with Inception-v2 \cite{BatchNorm}, Inception-v3 \cite{szegedy2016rethinking} and PolyNet \cite{zhang2016polynet}.
We show that this family of models achieve state-of-the-art performance with fewer floating point operations and parameters than comparable architectures. Second, we demonstrate that by adjusting the scale of the model we can achieve state-of-the-art performance at smaller computational budgets, exceeding streamlined CNNs hand-designed for this operating regime \cite{howard2017mobilenets, shufflenet}.

Note we do not have residual connections between convolutional cells as the models learn skip connections on their own. We empirically found manually inserting residual connections between cells to not help performance.
Our training setup on ImageNet is similar to \cite{szegedy2016rethinking}, but please see Appendix \ref{section:appendix-details} for details.

\begin{table*}[h!]
\centering
\small
\begin{tabular}{lc|cc|cc}
\toprule
\multicolumn{1}{c}{\bf Model} & {\bf image size} & \multicolumn{1}{l}{\bf \# parameters} & \bf Mult-Adds & \bf Top 1 Acc. (\%) & \bf Top 5 Acc. (\%) \\ \midrule
Inception V2~\cite{BatchNorm} & 224$\times$224 & 11.2\,M & 1.94\,B & 74.8 & 92.2 \\
\textbf{NASNet-A (5 @ 1538)} & \textbf{299$\times$299} & \textbf{10.9\,M} & \textbf{2.35\,B} & \textbf{78.6} & \textbf{94.2} \\
\midrule
Inception V3~\cite{szegedy2016rethinking} & 299$\times$299 & 23.8\,M & 5.72\,B & 78.8 & 94.4 \\
Xception~\cite{chollet2016xception}& 299$\times$299 & 22.8\,M & 8.38\,B & 79.0 & 94.5 \\
Inception ResNet V2~\cite{szegedy2016inception} & 299$\times$299 & 55.8\,M & 13.2\,B & 80.1 & 95.1 \\
\textbf{NASNet-A (7 @ 1920)} & \textbf{299$\times$299} & \textbf{22.6\,M} & \textbf{4.93\,B} & \textbf{80.8} & \textbf{95.3} \\
\midrule
ResNeXt-101 (64 x 4d)~\cite{xie2016aggregated} & 320$\times$320 & 83.6\,M & 31.5\,B & 80.9 & 95.6 \\
PolyNet~\cite{zhang2016polynet} & 331$\times$331 & 92\,M & 34.7\,B & 81.3 & 95.8 \\
DPN-131~\cite{dualpath} & 320$\times$320 & 79.5\,M & 32.0\,B & 81.5 & 95.8 \\
{\bf SENet~\cite{hu2017squeeze}}& {\bf 320$\times$320} & {\bf 145.8\,M} & {\bf 42.3\,B} & {\bf 82.7} & {\bf 96.2} \\

\textbf{NASNet-A (6 @ 4032)} & \textbf{331$\times$331} & \textbf{88.9\,M} & \textbf{23.8\,B} & \textbf{82.7} & \textbf{96.2} \\
\bottomrule
\end{tabular}

\vspace{0.2cm}
\caption{Performance of architecture search and other published state-of-the-art models on ImageNet classification. Mult-Adds indicate the number of composite multiply-accumulate operations for a single image. Note that the composite multiple-accumulate operations are calculated for the image size reported in the table. Model size for \cite{hu2017squeeze} calculated from open-source implementation.}
\label{tab:imagenet}
\end{table*}

\begin{table*}[h!]
\centering
\small
\begin{tabular}{l|cc|cc}
\toprule
\multicolumn{1}{c|}{\bf Model} & \multicolumn{1}{l}{\bf \# parameters} & \bf Mult-Adds & \bf Top 1 Acc. (\%) & \bf Top 5 Acc. (\%)  \\ \midrule
Inception V1~\cite{szegedy2015going} & 6.6M & 1,448\,M & $\;\;$69.8$\;^{\dagger}$ & 89.9 \\
MobileNet-224 \cite{howard2017mobilenets} & 4.2\,M & 569\,M & 70.6 & 89.5  \\
ShuffleNet (2x) \cite{shufflenet} & $\sim$ 5M & 524\,M & 70.9 & 89.8 \\
\midrule
\textbf{NASNet-A (4 @ 1056)} & \textbf{5.3\,M} & \textbf{564\,M} & \textbf{74.0} & \textbf{91.6} \\
NASNet-B (4 @ 1536) & 5.3M & 488\,M & 72.8 & 91.3  \\
NASNet-C (3 @ 960) & 4.9M & 558\,M & 72.5 & 91.0  \\
\bottomrule
\end{tabular}

\vspace{0.2cm}
\caption{Performance on ImageNet classification on a subset of models operating in a constrained computational setting, i.e., $< 1.5$\,B multiply-accumulate operations per image. All models use 224x224 images.
$\dagger$ indicates top-1 accuracy not reported in \cite{szegedy2015going} but from open-source implementation.}
\label{tab:imagenet-constrained}
\end{table*}





Table~\ref{tab:imagenet} shows that the convolutional cells discovered with CIFAR-10 generalize well to ImageNet problems.
In particular, each model based on the convolutional cells exceeds the predictive performance of the corresponding hand-designed model.
Importantly, the largest model achieves a new state-of-the-art performance for ImageNet (82.7\%) based on single, non-ensembled predictions, surpassing previous best published result by $\sim$1.2\% \cite{dualpath}. Among the unpublished works, our model is on par with the best reported result of 82.7\%~\cite{hu2017squeeze}, while having significantly fewer floating point operations.
Figure \ref{figure:flops_vs_accuracy} shows a complete summary of our results in comparison with other published results. Note the family of models based on convolutional cells provides an envelope over a broad class of human-invented architectures.

Finally, we test how well the best convolutional cells may perform in a resource-constrained setting, e.g.,  mobile devices (Table \ref{tab:imagenet-constrained}). In these settings, the number of floating point operations is severely constrained and predictive performance must be weighed against latency requirements on a device with limited computational resources. MobileNet \cite{howard2017mobilenets} and ShuffleNet \cite{shufflenet} provide state-of-the-art results obtaining 70.6\% and 70.9$\%$ accuracy, respectively on 224x224 images using $\sim$550M multliply-add operations. An architecture constructed from the best convolutional cells  achieves superior predictive performance (74.0\% accuracy) surpassing previous models but with comparable computational demand. In summary, we find that the learned convolutional cells are flexible across model scales achieving state-of-the-art performance across almost 2 orders of magnitude in computational budget.


\subsection{Improved features for object detection}

\begin{table*}[h!]
\centering
\small
\begin{tabular}{lc|cc}
\toprule
{\bf Model} & {\bf resolution} & {\bf mAP (mini-val)} & {\bf mAP (test-dev)} \\
\midrule
MobileNet-224 \cite{howard2017mobilenets} & $600\times600$ & 19.8\% & -  \\
ShuffleNet (2x) \cite{shufflenet} & $600\times600$  & 24.5\%$^\dagger$ & - \\ 
\textbf{NASNet-A (4 @ 1056)} & $600\times600$ & \textbf{29.6\%} & - \\
\midrule
ResNet-101-FPN \cite{lin2016feature} & 800 (short side) & - & 36.2\% \\
Inception-ResNet-v2 (G-RMI) \cite{huang2016speed} & $600\times600$ & 35.7\% & 35.6\% \\
Inception-ResNet-v2 (TDM) \cite{shrivastava2016beyond} & $600\times1000$ & 37.3\% & 36.8\% \\
\textbf{NASNet-A (6 @ 4032)} & $800\times800$ & 41.3\% & 40.7\% \\
\textbf{NASNet-A (6 @ 4032)} & $1200\times1200$ & \textbf{43.2\%} & {\bf 43.1\%} \\
\midrule
ResNet-101-FPN (RetinaNet) \cite{lin2017focal} & 800 (short side) & - & 39.1\%  \\
\bottomrule
\end{tabular}
\vspace{0.2cm}
\caption{Object detection performance on COCO on {\it mini-val} and {\it test-dev} datasets across a variety of image featurizations. All results are with the Faster-RCNN object detection framework \cite{faster_rcnn} from a single crop of an image. Top rows highlight mobile-optimized image featurizations, while bottom rows indicate computationally heavy image featurizations geared towards achieving best results. All {\it mini-val} results employ the same 8K subset of validation images in \cite{huang2016speed}.}
\label{table:mscoco}
\end{table*}

Image classification networks provide generic image features that may be transferred to other computer vision problems \cite{donahue2014decaf}. One of the most important problems is the spatial localization of objects within an image. To further validate the  performance of the family of NASNet-A networks, we test whether object detection systems derived from NASNet-A lead to improvements in object detection \cite{huang2016speed}.

To address this question, we plug in the family of NASNet-A networks pretrained on ImageNet into the Faster-RCNN object detection pipeline \cite{faster_rcnn} using an open-source software platform \cite{huang2016speed}. We retrain the resulting object detection pipeline on the combined COCO training plus validation dataset excluding 8,000 mini-validation images.
We perform single model evaluation using 300-500 RPN proposals per image. In other words, we only
pass a single image through a single network. We evaluate the model on the COCO {\it mini-val} \cite{huang2016speed} and {\it test-dev} dataset and report the mean average precision (mAP) as computed with the standard COCO metric library \cite{lin2014microsoft}. We perform a simple search over learning rate schedules to identify the best possible model. Finally, we examine the behavior of two object detection systems employing the best performing NASNet-A image featurization (NASNet-A, $6$ @ $4032$) as well as the image featurization geared towards mobile platforms (NASNet-A, $4$ @ $1056$).

For the mobile-optimized network, our resulting system achieves a mAP of 29.6\% -- exceeding previous mobile-optimized networks that employ Faster-RCNN by over 5.0\% (Table \ref{table:mscoco}). For the best NASNet network, our resulting network operating on images of the same spatial resolution (800 $\times$ 800) achieves mAP = 40.7\%, exceeding equivalent object detection systems based off lesser performing image featurization (i.e. Inception-ResNet-v2) by 4.0\% \cite{huang2016speed,shrivastava2016beyond} (see Appendix for example detections on images and side-by-side comparisons). Finally, increasing the spatial resolution of the input image results in the best reported, single model result for object detection of 43.1\%, surpassing the best previous best by over 4.0\% \cite{lin2017focal}.\footnote{A primary advance in the best reported object detection system is the introduction of a novel loss \cite{lin2017focal}. Pairing this loss with NASNet-A image featurization may lead to even further performance gains. Additionally, performance gains are achievable through ensembling multiple inferences across multiple model instances and image crops (e.g., \cite{huang2016speed}).} These results provide further evidence that NASNet provides superior, generic image features that may be transferred across other computer vision tasks. Figure~\ref{figure:object-detection} and Figure~\ref{figure:object-detection-examples} in Appendix~\ref{sec:object_detection} show four examples of object detection results produced by NASNet-A with the Faster-RCNN framework.


\subsection{Efficiency of architecture search methods}
\label{sec:random_search}

\begin{figure}[h!]
\begin{center}
\includegraphics[width=\columnwidth]{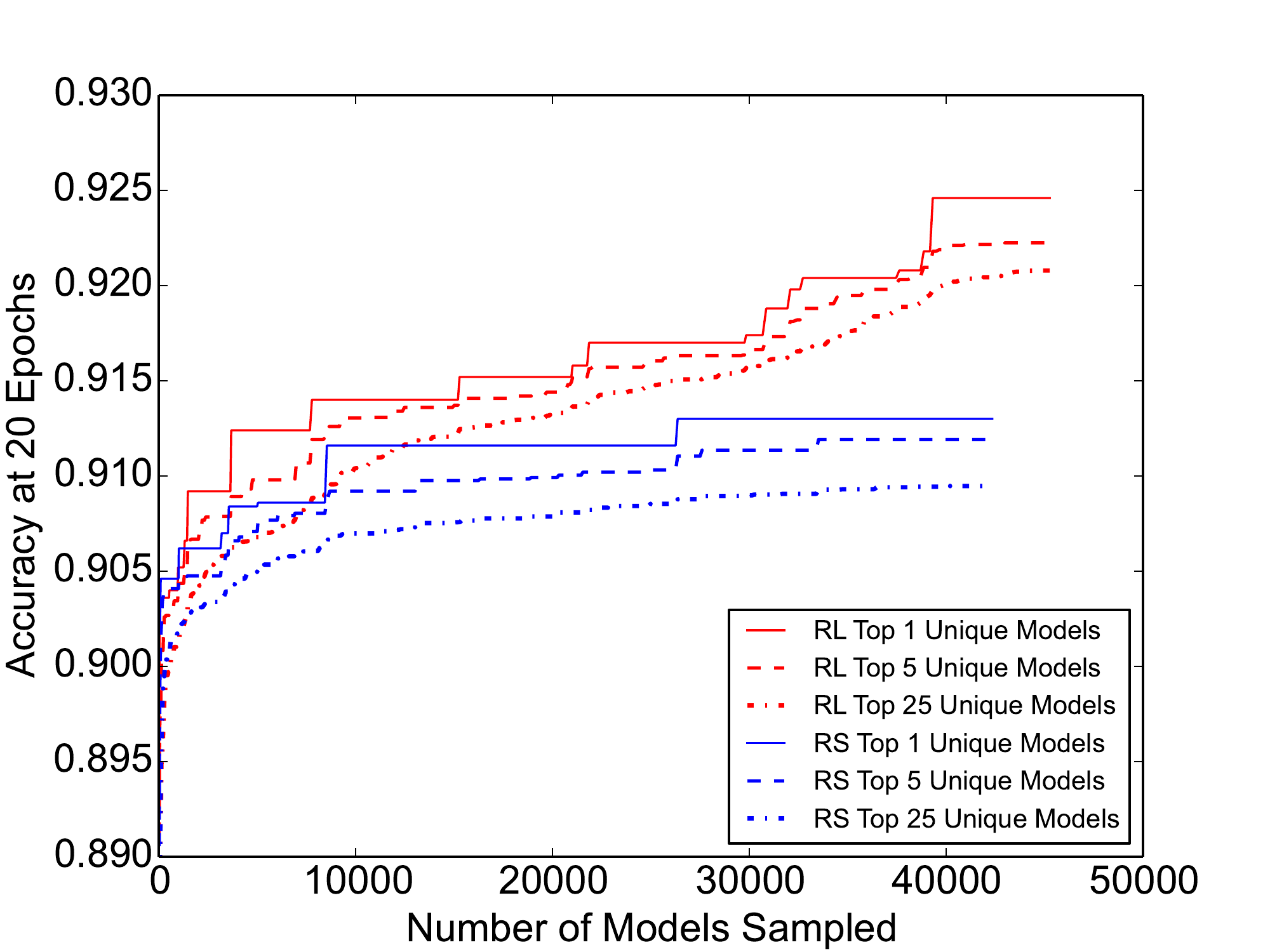}
\caption{Comparing the efficiency of random search (RS) to reinforcement learning (RL) for learning neural architectures. The x-axis measures the total number of model architectures sampled, and the y-axis is the validation performance on CIFAR-10 after 20 epochs of training. 
}
\label{figure:random-search}
\end{center}
\end{figure}

Though what search method to use is not the focus of the paper, an open question is how effective is the reinforcement learning search method. In this section, we study the effectiveness of reinforcement learning for architecture search on the CIFAR-10 image classification problem and compare it to brute-force random search (considered to be a very strong baseline for black-box optimization~\cite{bergstra2012random}) given an equivalent amount of computational resources. 


Figure \ref{figure:random-search} shows the performance of reinforcement learning (RL) and random search (RS) as more model architectures are sampled. Note that the best model identified with RL is significantly better than the best model found by RS by over 1\% as measured by on CIFAR-10. Additionally, RL finds an entire range of models that are of superior quality to random search. We observe this in the mean performance of the top-5 and top-25 models identified in RL versus RS.
We take these results to indicate that although RS may provide a viable search strategy, RL finds better architectures in the NASNet search space.


%% file: conclusions.tex
\section{Conclusion}
In this work, we demonstrate how to learn scalable, convolutional cells from data that transfer to multiple image classification tasks. The learned architecture is quite flexible as it may be scaled in terms of computational cost and parameters to easily address a variety of problems. In all cases, the accuracy of the resulting model exceeds all human-designed models -- ranging from models designed for mobile applications to computationally-heavy models designed to achieve the most accurate results.

The key insight in our approach is to design a search space that decouples the complexity of an architecture from the depth of a network. This resulting search space permits identifying good architectures on a small dataset (i.e., CIFAR-10) and transferring the learned architecture to image classifications across a range of data and computational scales.

The resulting architectures approach or exceed state-of-the-art performance in both CIFAR-10 and ImageNet datasets with less computational demand than human-designed architectures \cite{szegedy2016rethinking,BatchNorm,zhang2016polynet}. The ImageNet results are particularly important because many state-of-the-art computer vision problems (e.g.,  object detection \cite{huang2016speed}, face detection \cite{schroff2015facenet}, image localization \cite{weyand2016planet}) derive image features or architectures from ImageNet classification models. For instance, we find that image features obtained from ImageNet used in combination with the Faster-RCNN framework achieves state-of-the-art object detection results. Finally, we demonstrate that we can use the resulting learned architecture to perform ImageNet classification with reduced computational budgets that outperform streamlined architectures targeted to mobile and embedded platforms \cite{howard2017mobilenets, shufflenet}.

%% file: appendix.tex
\newpage
\appendix
\section*{Appendix}
\section{Experimental Details}
\label{section:appendix-details}

\subsection{Dataset for Architecture Search}
The CIFAR-10 dataset~\cite{krizhevsky2009learning} consists of 60,000 32x32 RGB images across 10 classes (50,000 train and 10,000 test images).
We partition a random subset of 5,000 images from the training set to use as a validation set for the controller RNN. All images are whitened and then undergone several data augmentation steps: we randomly crop 32x32 patches from upsampled images of size 40x40 and apply random horizontal flips. This data augmentation procedure is common among related work.


\subsection{Controller architecture}
The controller RNN is a one-layer LSTM~\cite{lstm} with 100 hidden units at each layer and $2 \times 5B$ softmax predictions for the two convolutional cells (where $B$ is typically 5) associated with each architecture decision. Each of the $10B$ predictions of the controller RNN is associated with a probability. The joint probability of a child network is the product of all probabilities at these $10B$ softmaxes. This joint probability is used to compute the gradient for the controller RNN. The gradient is scaled by the validation accuracy of the child network to update the controller RNN such that the controller assigns low probabilities for bad child networks and high probabilities for good child networks. 

Unlike~\cite{zoph2017neural}, who used the REINFORCE rule~\cite{Williams92simplestatistical} to update the controller, we employ Proximal Policy Optimization (PPO)~\cite{SchulmanWDRK17} with learning rate 0.00035 because training with PPO is faster and more stable. To encourage exploration we also use an entropy penalty with a weight of 0.00001. In our implementation, the baseline function is an exponential moving average of previous rewards with a weight of 0.95. The weights of the controller are initialized uniformly between -0.1 and 0.1.

\subsection{Training of the Controller}
For distributed training, we use a workqueue system where all the samples generated from the controller RNN are added to a global workqueue. A free ``child" worker in a distributed worker pool asks the controller for new work from the global workqueue. Once the training of the child network is complete, the accuracy on a held-out validation set is computed and reported to the controller RNN. In our experiments we use a child worker pool size of 450, which means there are 450 networks being trained on 450 GPUs concurrently at any time. Upon receiving enough child model training results, the controller RNN will perform a gradient update on its weights using PPO and then sample another batch of architectures that go into the global workqueue. This process continues until a predetermined number of architectures have been sampled. In our experiments, this predetermined number of architectures is 20,000 which means the search process is terminated after 20,000 child models have been trained. Additionally, we update the controller RNN with minibatches of 20 architectures. Once the search is over, the top 250 architectures are then chosen to train until convergence on CIFAR-10 to determine the very best architecture.

\subsection{Details of architecture search space}
\begin{figure*}[t]
\begin{center}
\includegraphics[width=1.9\columnwidth]{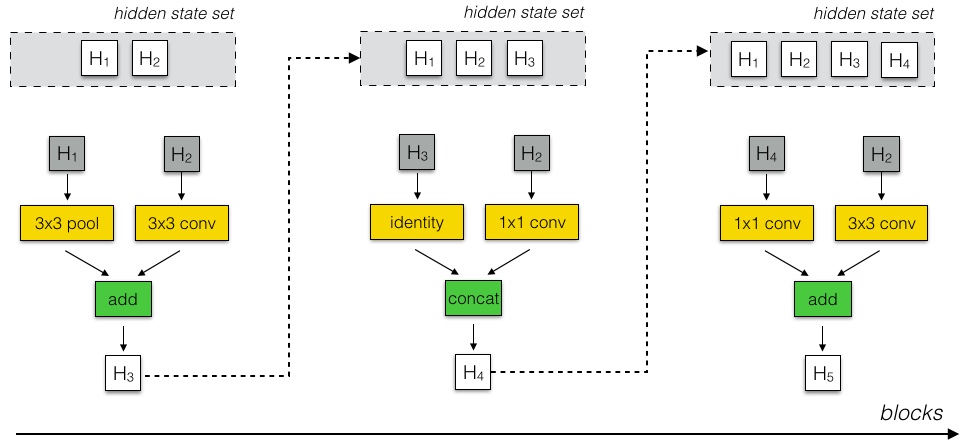}
\caption{Schematic diagram of the NASNet search space. Network motifs are constructed recursively in stages termed blocks. Each block consists of the controller selecting a pair of hidden states (dark gray), operations to perform on those hidden states (yellow) and a combination operation (green). The resulting hidden state is retained in the set of potential hidden states to be selected on subsequent blocks.}
\label{figure:nas-search-space}
\end{center}
\end{figure*}

We performed preliminary experiments to identify a flexible, expressive search space for neural architectures that learn effectively. Generally, our strategy for preliminary experiments involved small-scale explorations to identify how to run large-scale architecture search.

\begin{itemize}
\item All convolutions employ ReLU nonlinearity. Experiments with ELU nonlinearity \cite{clevert2015fast} showed minimal benefit.
\item To ensure that the shapes always match in convolutional cells, 1x1 convolutions are inserted as necessary.
\item Unlike \cite{howard2017mobilenets}, all depthwise separable convolution do not employ Batch Normalization and/or a ReLU between the depthwise and pointwise operations.
\item All convolutions followed an ordering of ReLU, convolution operation and Batch Normalization following \cite{identity-mappings}.
\item Whenever a separable convolution is selected as an operation by the model architecture, the separable convolution is applied twice to the hidden state. We found this empirically to improve overall performance.
\end{itemize}

\subsection{Training with ScheduledDropPath}
We performed several experiments with various stochastic regularization methods. Naively applying dropout \cite{srivastava2014dropout} across convolutional filters degraded performance. However,  we discovered a new technique called \emph{ScheduledDropPath}, a modified version of DropPath~\cite{larsson2016fractalnet}, that works well in regularizing NASNets. In DropPath, we stochastically drop out each path (i.e., edge with a yellow box in Figure \ref{figure:cell_structure}) in the cell with some fixed probability. This is similar to \cite{stochastic} and \cite{zhang2016polynet} where they dropout full parts of their model during training and then at test time scale the path by the probability of keeping that path during training. Interestingly we also found that DropPath alone does not help NASNet training much, but DropPath with \emph{linearly increasing the probability of dropping out a path over the course of training}  significantly improves the final performance for both CIFAR and ImageNet experiments. We name this method ScheduledDropPath.

\subsection{Training of CIFAR models}
All of our CIFAR models use a single period cosine decay as in \cite{loshchilov2016sgdr, gastaldi17shakeshake}.  All models use the momentum optimizer with momentum rate set to 0.9. All models also use L2 weight decay. Each architecture is trained for a fixed 20 epochs on CIFAR-10 during the architecture search process. Additionally, we found it beneficial to use the cosine learning rate decay during the 20 epochs the CIFAR models were trained as this helped to further differentiate good architectures. We also found that having the CIFAR models use a small $N=2$ during the architecture search process allowed for models to train quite quickly, while still finding cells that work well once more were stacked.

\subsection{Training of ImageNet models}

We use ImageNet 2012 ILSVRC challenge data for large scale image classification. The dataset consists of $\sim$ 1.2M images labeled across 1,000 classes \cite{deng2009imagenet}. Overall our training and testing procedures are almost identical to \cite{szegedy2016rethinking}. ImageNet models are trained and evaluated on 299x299 or 331x331 images using the same data augmentation procedures as described previously \cite{szegedy2016rethinking}. We use distributed synchronous SGD to train the ImageNet model with 50 workers (and 3 backup workers) each with a Tesla K40 GPU \cite{synchsgd}. We use RMSProp with a decay of 0.9 and epsilon of 1.0. Evaluations are calculated using with a running average of parameters over time with a decay rate of 0.9999. We use label smoothing with a value of 0.1 for all ImageNet models as done in \cite{szegedy2016rethinking}. Additionally, all models use an auxiliary classifier located at 2/3 of the way up the network. The loss of the auxiliary classifier is weighted by 0.4 as done in \cite{szegedy2016rethinking}. We empirically found our network to be insensitive to the number of parameters associated with this auxiliary classifier along with the weight associated with the loss. All models also use L2 regularization.  The learning rate decay scheme is the exponential decay scheme used in \cite{szegedy2016rethinking}.  Dropout is applied to the final softmax matrix with probability 0.5.


\section{Additional Experiments}
\label{sec:othernasnet}
\begin{figure}[h!]
\begin{center}
\includegraphics[width=\columnwidth]{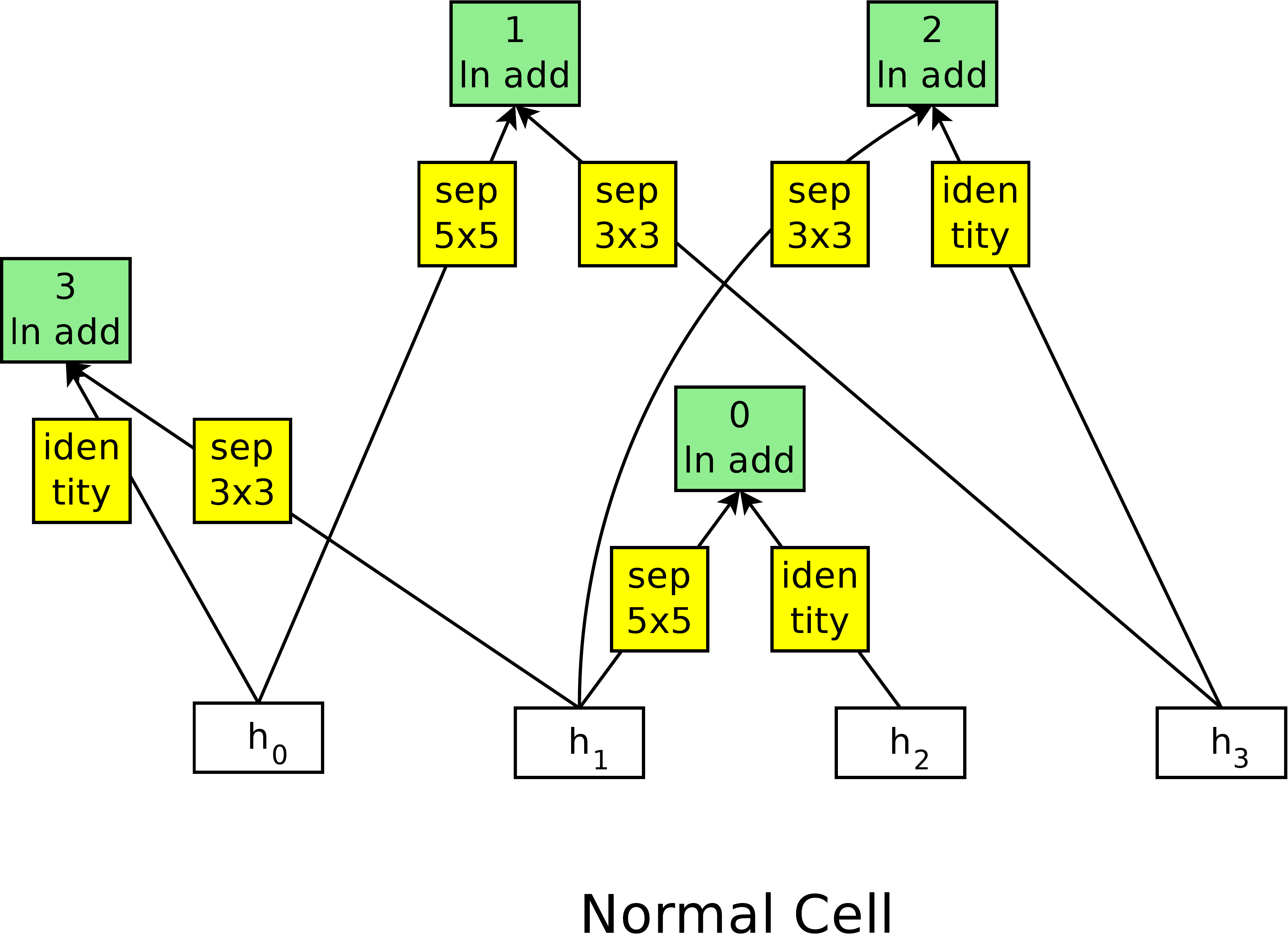} \\
\vspace{0.4cm}
\includegraphics[width=\columnwidth]{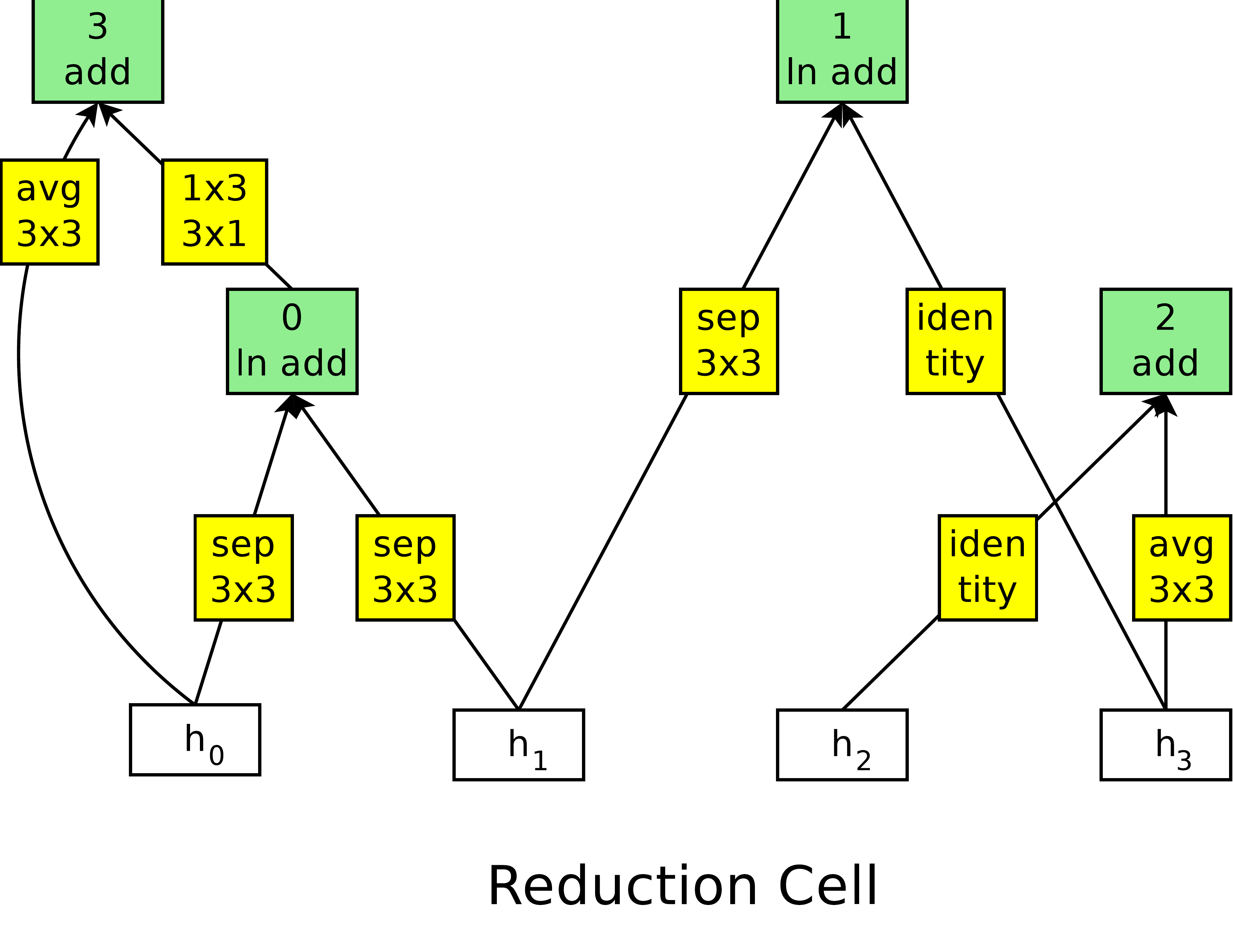}
\caption{Architecture of NASNet-B convolutional cell with $B=4$ blocks identified with CIFAR-10. The input (white) is the hidden state from previous activations (or input image).
Each convolutional cell is the result of $B$ blocks.
A single block is corresponds to two primitive operations (yellow) and a combination operation (green). As do we not concatenate the output hidden states, each output hidden state is used as a hidden state in the future layers. Each cell takes in 4 hidden states and thus needs to also create 4 output hidden states. Each output hidden state is therefore labeled with 0, 1, 2, 3 to represent the next four layers in that order.
}
\label{figure:cell_structure_b}
\end{center}
\end{figure}

We now present two additional cells that performed well on CIFAR and ImageNet. The search spaces used for these cells are slightly different than what was used for NASNet-A. For the NASNet-B model in Figure \ref{figure:cell_structure_b} we do not concatenate all of the unused hidden states generated in the convolutional cell. Instead all of the hiddenstates created within the convolutional cell, even if they are currently used, are fed into the next layer. Note that $B=4$ and there are 4 hiddenstates as input to the cell as these numbers must match for this cell to be valid. We also allow addition followed by layer normalization \cite{ba2016layer} or instance normalization \cite{ulyanov2016instance} to be predicted as two of the combination operations within the cell, along with addition or concatenation.

\begin{figure}[h!]
\begin{center}
\includegraphics[width=\columnwidth]{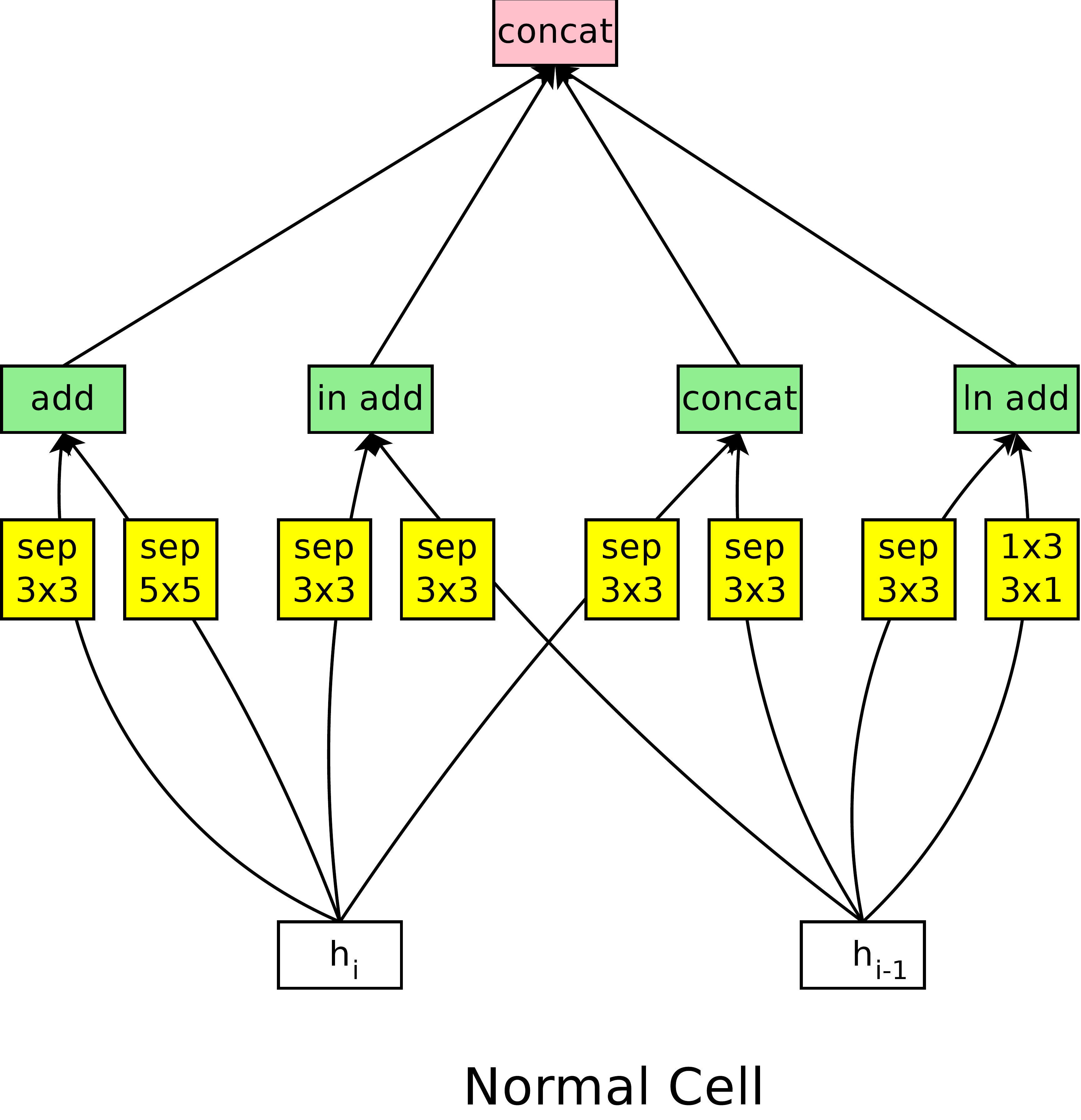} \\
\vspace{0.4cm}
\includegraphics[width=\columnwidth]{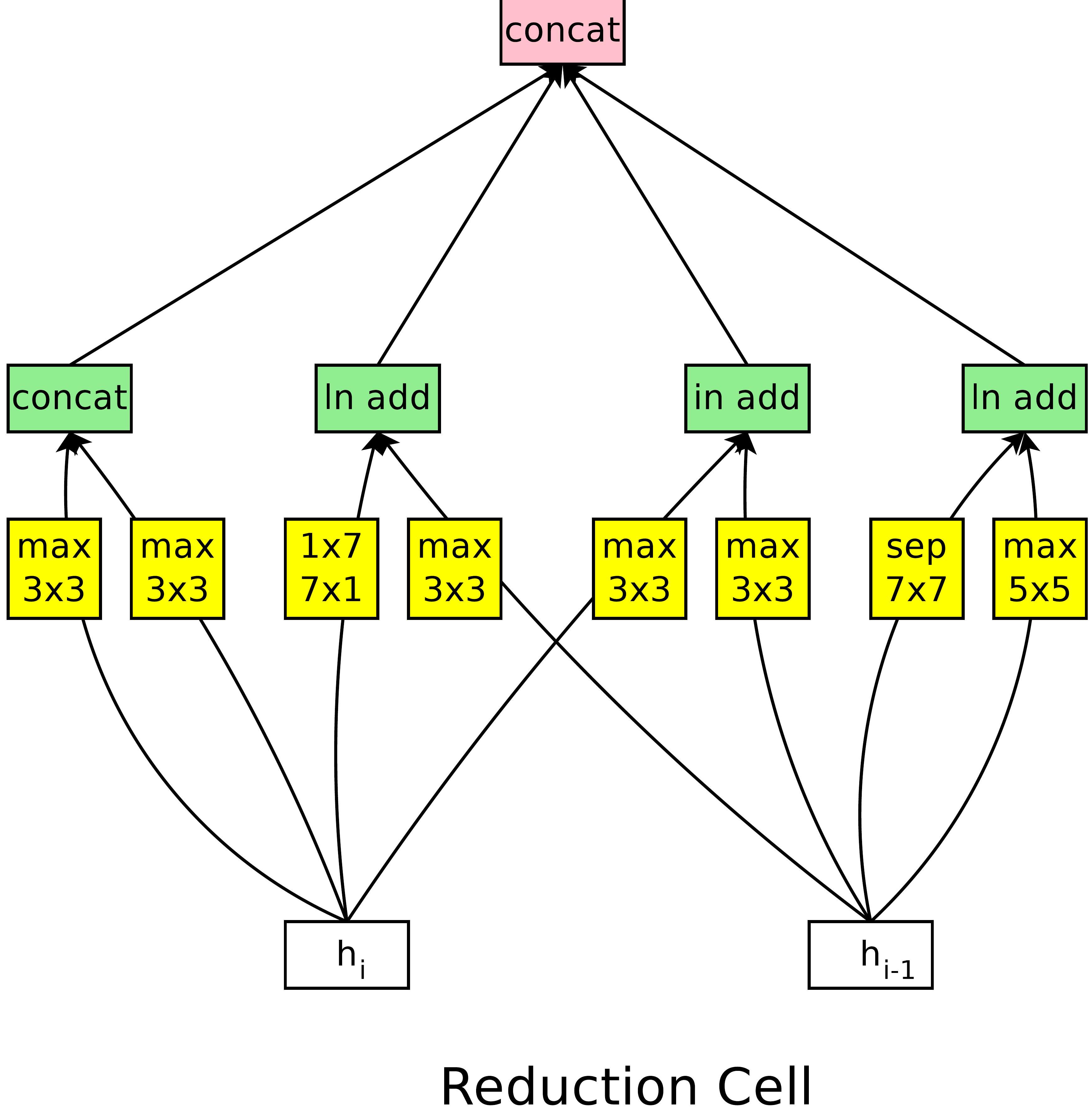}
\caption{Architecture of NASNet-C convolutional cell with $B=4$ blocks identified with CIFAR-10. The input (white) is the hidden state from previous activations (or input image). The output (pink) is the result of a concatenation operation across all resulting branches.
Each convolutional cell is the result of $B$ blocks.
A single block corresponds to two primitive operations (yellow) and a combination operation (green). 
}
\label{figure:cell_structure_c}
\end{center}
\end{figure}

For NASNet-C (Figure~\ref{figure:cell_structure_c}), we concatenate all of the unused hidden states generated in the convolutional cell like in NASNet-A, but now we allow the prediction of addition followed by layer normalization or instance normalization like in NASNet-B.

\section{Example object detection results}
\label{sec:object_detection}
\begin{figure}[h!]
\begin{center}
\includegraphics[width=0.98\columnwidth]{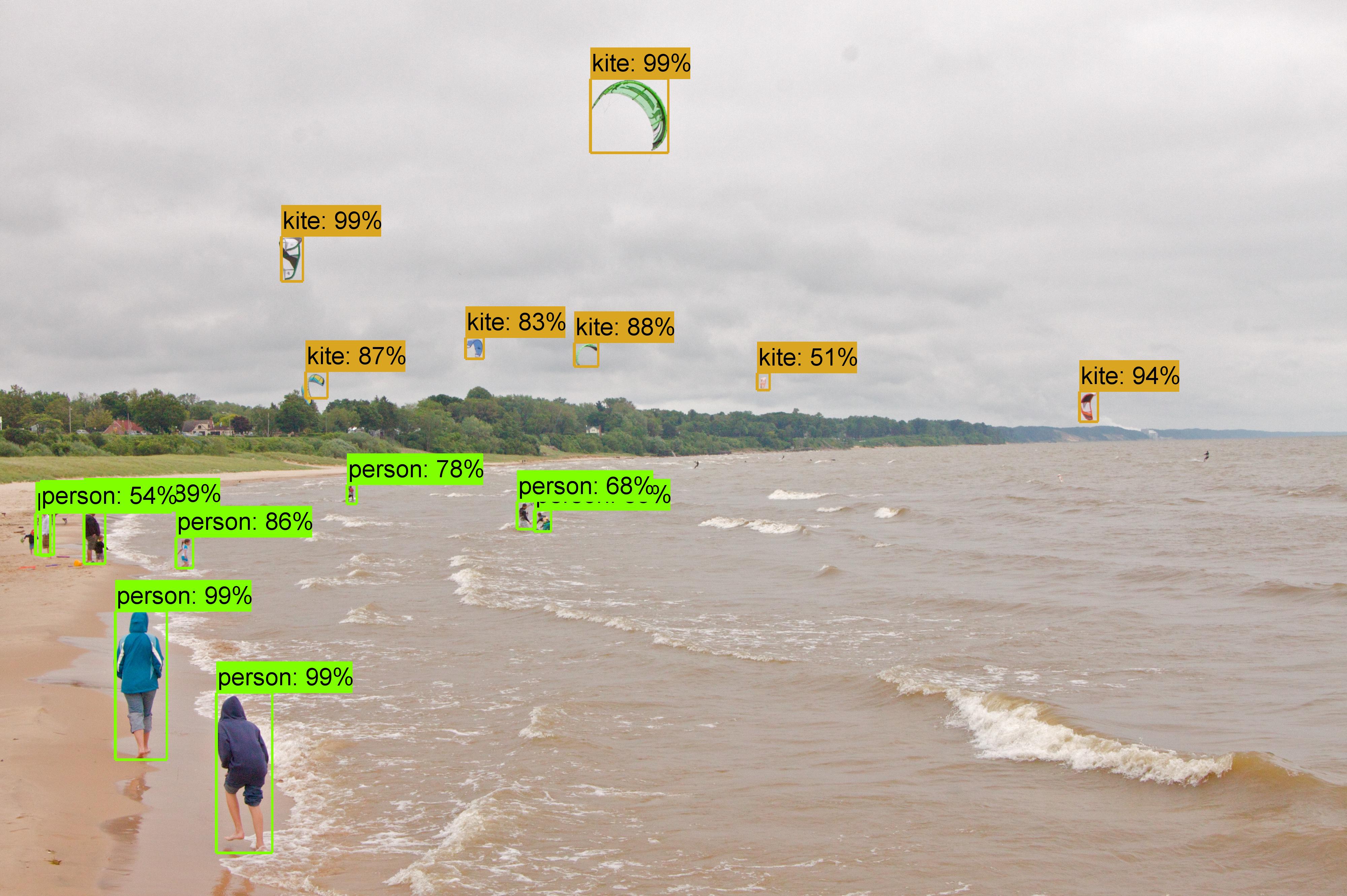}
\hfill
\includegraphics[width=0.98\columnwidth]{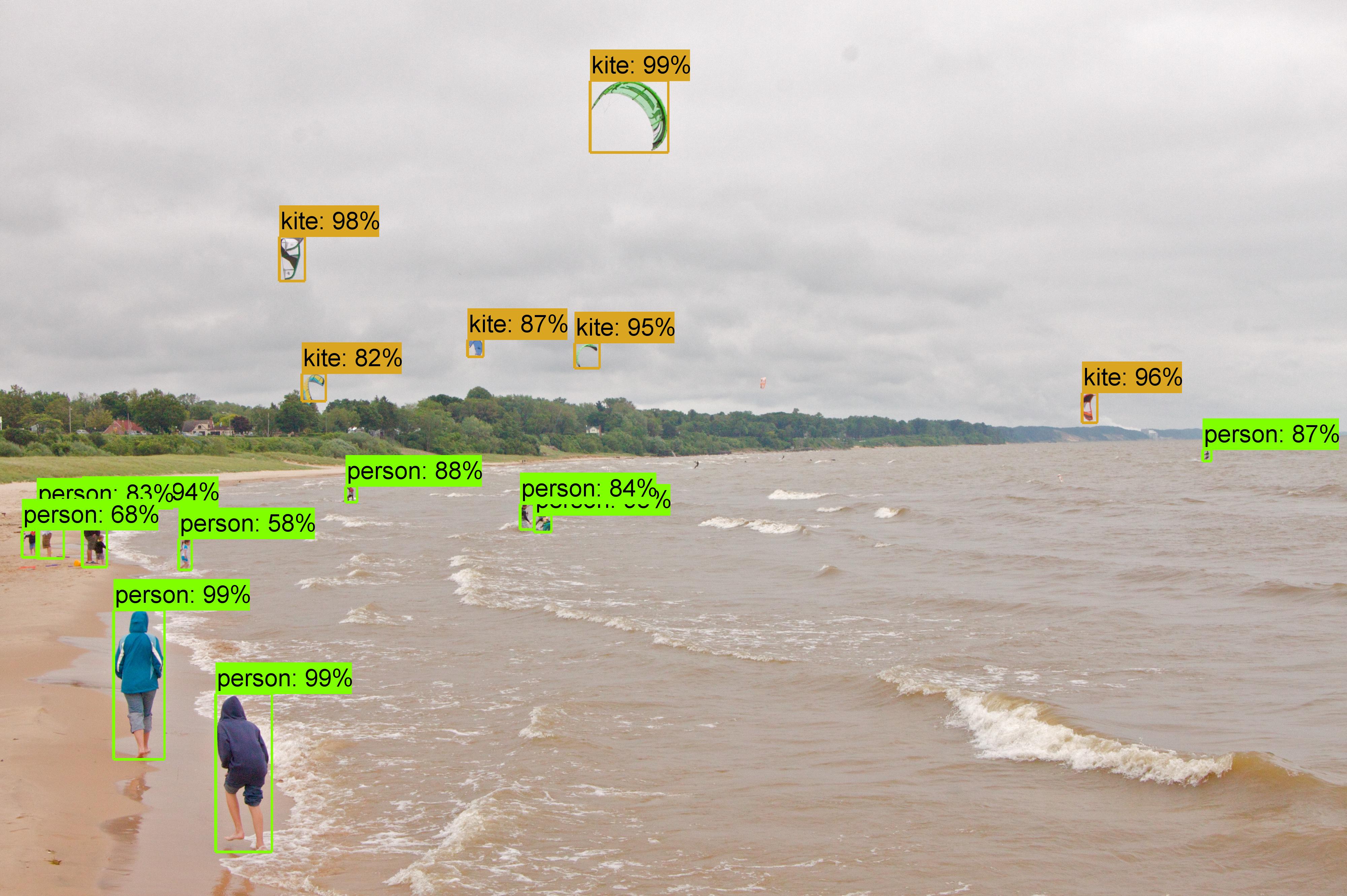}
\caption{Example detections showing improvements of object detection over previous state-of-the-art model for Faster-RCNN with Inception-ResNet-v2 featurization \cite{huang2016speed} (top)
and NASNet-A featurization (bottom). 
}
\label{figure:object-detection}
\end{center}
\end{figure}

\begin{figure}[h!]
\begin{center}
\includegraphics[width=\columnwidth]{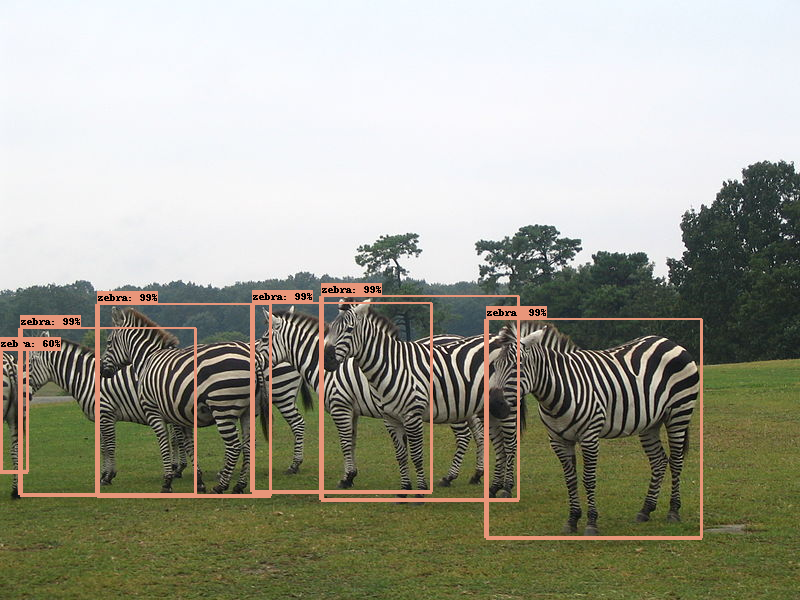}
\hfill
\includegraphics[width=\columnwidth]{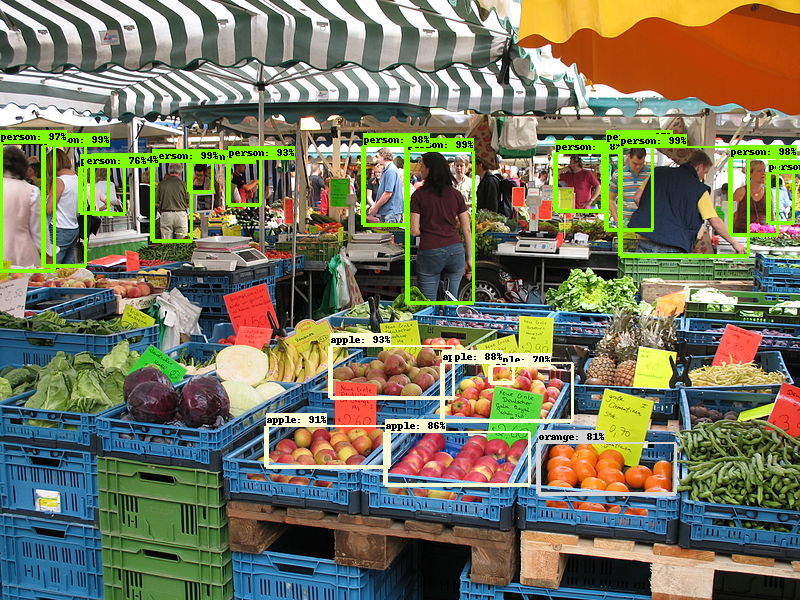}
\hfill
\includegraphics[width=\columnwidth]{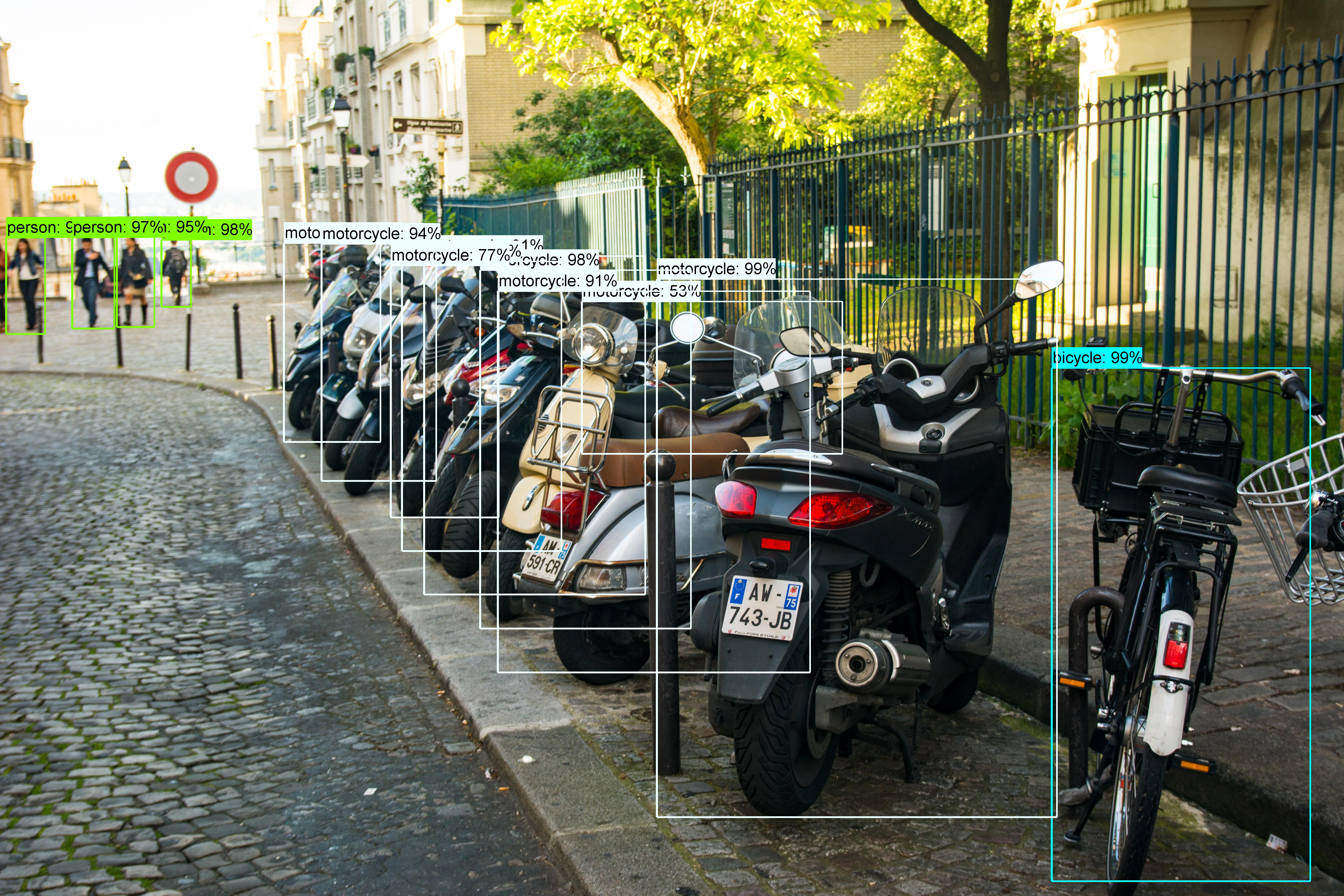}
\caption{Example detections of best performing NASNet-A featurization with Faster-RCNN trained on COCO dataset. Top and middle images courtesy of \texttt{http://wikipedia.org}. Bottom image courtesy of Jonathan Huang}
\label{figure:object-detection-examples}
\end{center}
\end{figure}

Finally, we will present examples of object detection results on the COCO dataset in  Figure~\ref{figure:object-detection} and Figure~\ref{figure:object-detection-examples}. As can be seen from the figures, NASNet-A featurization works well with Faster-RCNN and gives accurate localization of objects.

